# 1st Place Solution in Google Universal Images Embedding

Shihao Shao*#
Peking University
Health Science Center
shaoshihao@pku.edu.cn

Qinghua Cui*#
Peking University
Health Science Center
cuiqinghua@bjmu.edu.cn

## Abstract

*This paper presents the 1st place solution for the Google Universal Images Embedding Competition on Kaggle. The highlighted part of our solution is based on 1) A novel way to conduct training and fine-tuning; 2) The idea of a better ensemble in the pool of models that make embedding; 3) The potential trade-off between fine-tuning on high-resolution and overlapping patches; 4) The potential factors to work for the dynamic margin. Our solution reaches 0.728 in the private leader board, which achieve 1st place in Google Universal Images Embedding Competition.*

## 1. Introduction

Google Universal Image Embedding Competition [1] is a part of the Instance-Level Recognition workshop(ILR) hosted on Kaggle. 1,022 teams join the competition this year. The competition is based on the growing need to extract embedding from images, leveraging the power of image representation. The image to be evaluated in this competition covers several types: apparel & accessories, packaged goods, landmarks, furniture & home decor, storefronts, dishes, artwork, toys, memes, illustrations, and cars. This Competition requires contestants to develop a model with the capacity to generate a 64-d embedding for each image. Then the back-end server will retrieve the image of the same instance via searching based on KNN(k = 5). Each submission is evaluated according to the mean Precision @ 5 metrics and should be within a total 9 hours running limit.

## 2. Configuration for the Best Performance

For datasets selecting, we finally cover Products-10K [2], Shopee [3], MET Artwork Dataset [4], Alibaba goods [5], H&M Personalized Fashion [6], GPR1200 [7], GLDv2-Full [8], DeepFashion - Consumer-to-shop Clothes Retrieval Benchmark part [9]. Concretely, Products-10K is used in the last fine-tuning part, and the rest is for the first-stage training. We adapt weights of VIT-H pre-trained on Laion-2B, a subset of Laion-5B [10], as our baseline model. To meet the requirement of the embedding length, we add a linear projection layer to squeeze the embedding into 64-d. Then, an ArcFace [11] head was adapted together with this modification. A Dropout layer is inserted between the last and the second-to-last linear layer with a drop rate of 0.2. We use SGD with momentum as our optimizer, and an L2 weights decay rate of 1.5e-4.

The training and fine-tuning part are different from the common paradigm. Here we have two main ways to better fine-tune pre-trained weights in new domains. 1) LP-FT [12], to fine-tune the last layer, and unfreeze all layers to continue fine-tuning with a lower learning rate; 2) Set the fully-connected layer and the backbone with different learning rates to train (not sure who proposed first). We propose a novel way to make the model better converged and avoid over-fitting. We trained on the last 2 fully-connected layers to completely converge for 6 epochs since there is less risky to cause over-fitting for only training on parameters of two layers. Then, instead of LP-FT, we freeze the linear layer and only train on the backbone part, for 3 epochs. The reason why we did not use methods 1) and 2) is written in the Method part below. The whole training process will be conducted on Products-10K again after training on the rest of the datasets.

Empirically, fine-tuning higher resolutions will further improve performance. After doing the above training, we fine-tuned the model on 280*280, and then 290*290 but with 4 pixels overlapping in patches. We have training on whole datasets again with these higher resolutions but with half epochs and learning rates. We did not use any Test-Time Augmentation in the inference. The final score is 0.732 and 0.728 in public and private leader boards, respectively.

## 3. Methods

In this section, we will discuss the several idea come up in this competition. Some of them has already some experimental basis, and the others remain assumption.

\* contributes equally to this work
\# are the corresponding authors of this paper

## 3.1. Fine-tuning with Central-Vector Orientation Locking

Fine-tuning the last layer is a common approach to fine-tune models with pre-trained weights for long. Training on all blocks of model will result in bad performance, since the random initial linear head harms the backbones with pre-trained weights. Recently, researchers have proposed novel fine-tuning paradigm. The first one, claiming it is better to fine-tune model with different learning rate adapting to final linear layers and backbones, respectively. The newest research is LP-FT, claiming a proper way to fine-tune all layers after training the last layer, which can have much more capacity regarding the power of generalization.

In practice, we find it easy to over-fit when we unfreeze the backbone part for further fine-tuning. In our experiments, we find the weights of the last layer shakes rapidly when we train on all the layers. Recall the last layer as a mathematic operation, it is:

$$\hat{y} = f_w(g_{w'}(x)) \qquad (1)$$

Where $f_w(*)$ denotes the last linear layer of the model with parameters $w$. $g_{w'}(*)$ represents the backbone part with parameters $w'$. $\hat{y}$ is the estimation of the embedding vector. $x$ is a given image as the input.

Here, $w$ is in the shape of $(E, C)$, $w_{i,j}$ is the i-th dimension of the embedding of the j-th class. So, $w_{*,j}$ is the embedding of the j-th class. The rapid shaking of $w_{*,j}$ indicates the changing of central orientation of each class. Since the central embedding already converged when we fine-tune the last layer, we assume that the central orientation is already well learned at that stage. The rapid shaking begins with the training on all layers, so this shaking may be a fitting to the training datasets at the cost of distorting the central orientation of each class embedding.

Here should be a way to overcome this issue, simply freezing the last layer while training on the rest part. We conducted an ablation study shown in Table 1.

| Linear Layer | Backbone | Public LB | Private LB |
|---|---|---|---|
| Freeze | UnFreeze | 0.680 | 0.678 |
| Unfreeze* | Unfreeze* | 0.678 | 0.677 |

Table 1. Ablation studies for Fine-tuning with Central-Vector Orientation Locking. * follows LT-FT paradigm to fine-tune.

## 3.2. Ensemble of embedding models

Model ensemble is a well-know technique to achieve better performance by fusion the different models in some ways. The most common approach is to simply calculate the weighted average of the output of each model. Surprisingly, model ensemble by averaging output does not give a superior result. It results in the worst performance among the models pool instead. Man of the year, in the forum of this competition, has conducted some reasonable analysis about this phenomenon. He supposes each model can determine different vital neurons for a class. The naive average of each model can result in confusing result. To extend this phenomenons to a more generalized form, we have an assumption: the ensemble works only if it meets the requirements below:

When we have,

$$\hat{y}_1 = f_{w_1}(g_{w'_1}(x)) \qquad (2)$$
$$\hat{y}_2 = f_{w_2}(g_{w'_2}(x)) \qquad (3)$$

The averaging of $\hat{y}_1$ and $\hat{y}_2$ works, only if:

$$\hat{y}_1 = k \times \hat{y}_2 \qquad (4)$$

Furthermore, if we train two models on the same datasets, due to the noise of randomly mini-batch selection and possibilities in augmentation, (4) can not be satisfied, so the ensemble does not work as it is in the most cases. $\hat{y}_*$ is determined by the central embedding of each class. Regarding this, we assume the distribution of $\hat{y}_*$ highly relies on the last layer of the model, since which is the definitions of the central orientation of each class. Therefore, in our VIT-L experiments, we use different hyper-parameters to fine-tune the backbone, leaving the last 2 fully-connected layer frozen, to conduct the ensemble. The results show it completely solve the issue in Table 2.

| Ensemble of the models | Public LB | Private LB |
|---|---|---|
| Last Layer freezing | 0.681 | 0.678 |
| Fine-tuning from the beginning | 0.669 | 0.668 |
| The best model only | 0.671 | 0.672 |

Table 2. Ablation studies for the ensemble methods

Besides this, we also assume different models, if only trained on the same datasets, will have their $\hat{y}_*$ can be transferred to each other via a linear operation.

For ∀i, j ∈ [1,2,3,…,N], where N is the amount of models for ensemble, ∃k, b ∈ R, we have:

$$\hat{y}_i = k \times \hat{y}_j + b$$

It means we can add a linear layer to a given model producing $\hat{y}_j$, making it close to $\hat{y}_i$, with the same performance maintained. In practice, we choose a model $f_{w_i}(g_{w'_i}(*))$ with the best performance. For any given other model $f_{w_j}(g_{w'_j}(*))$, we adapt a linear layer between $f_{w_j}(*)$ and $g_{w_j}(*)$, and replace $f_{w_j}(*)$ with $f_{w_i}(*)$ to make central embedding the same, which means we have $f_{w_i}(t_{w''_j}(g_{w'_j}(*)))$. During training, we need to hold $w_i$, $w'_j$ unchanged, and only fine-tune $w''_j$. We did not conduct ablation studies on this part, and decide to dive into it in future researches.

### 3.3. The potential trade-off between fine-tuning on high-resolution and overlapping patches

It is reported that the overlapping patches can help image segmentation for Vision Transformer models[13]. We found it also work in image embedding extraction. One of common approaches to gain better performance through fine-tuning in high resolutions. But it will gradually converge to certain performance, losing the trade-off between resolution and performance. We found that when high-resolution can not help in improving performance, adding overlapping between patches will keep making performance better. After we use the 336*336 resolution to fine-tune our model, it seems hard to improve in this stage. Instead, we use 290*290 resolution with 4 pixels patch overlapping, and get remarkable result shown in Table 3.

| Resolution/Overlapping | Public LB | Private LB |
|---|---|---|
| 336*336 | 0.732 | 0.728 |
| 290*290 | 0.728 | 0.728 |

Table 3. The resolution and overlapping that help the performance.

Many works like [14], has explored the trade-off between resolution, depth, width. We suppose the overlapping pixel can be put into the trade-off, which is what we want to further dive in.

### 3.4. The potential factors to work for the dynamic margin

[15] has proposed dynamic margin working better in the instance-level tasks. The insight from that is assuming the margin of ArcFace should be cooperated with the size of each class. However, dynamic margin does not perform better with vanilla ArcFace. We assume dynamic margin might need to consider the amount of classes as well, since dynamic margin is developed based on GLD-v2 datasets, which is far smaller than our case. Also, the limit size of embedding space is a constraint for classification on large amount of classes.

### 4. Summary

In this paper, we present our 1st place solution of Google Universal Images Embedding, and many other ideas come up with in this competition. Some of them were already conducted ablation studies, the rest of which remains us to discover. We achieve 0.732 and 0.728 in the public and private leader boards in this competition, respectively, which gets the 1st place in both 2 leader boards.